\documentclass{article} % For LaTeX2e
\usepackage{iclr2016_conference,times}
\usepackage{hyperref}
\usepackage{url}
\usepackage{amsmath}
\usepackage{multirow}
\usepackage{lipsum}
\usepackage{subfig}
\usepackage[pdftex]{graphicx}
\DeclareGraphicsExtensions{.pdf}
\title{Resiliency of Deep Neural Networks \\under Quantization}

\author{Wonyong~Sung, Sungho~Shin \& Kyuyeon~Hwang \\
Department of Electrical and Computer Engineering\\
Seoul National University\\
Seoul, 08826 Korea\\
\texttt{wysung@snu.ac.kr} \\
\texttt{shshin@dsp.snu.ac.kr} \\
\texttt{kyuyeon.hwang@gmail.com}\\
}

\begin{document}

\maketitle

\begin{abstract}
The complexity of deep neural network algorithms for hardware implementation can be much lowered by optimizing the word-length of weights and signals. 
Direct quantization of floating-point weights, however, does not show good performance when the number of bits assigned is small.  
Retraining of quantized networks has been developed to relieve this problem.  
In this work, the effects of quantization are analyzed for a feedforward deep neural network (FFDNN) and a convolutional neural network (CNN) when their network complexity is changed.   
The complexity of the FFDNN is controlled by varying the unit size in each hidden layer and the number of layers, while that of the CNN is done by modifying the feature map configuration. 
We find that some performance gap exists between the floating-point and the retrain-based ternary (+1, 0, -1) weight neural networks when the size is not large enough, but the discrepancy almost vanishes in fully complex networks whose capability is limited by the training data, rather than by the number of connections.  
This research shows that highly complex DNNs have the capability of absorbing the effects of severe weight quantization through retraining, but connection limited networks are less resilient. This paper also presents the effective compression ratio to guide the trade-off between the network size and the precision when the hardware resource is limited.
\end{abstract}

\section{Introduction}
Deep neural networks (DNNs) begin to find many real-time applications, such as speech recognition, autonomous driving, gesture recognition, and robotic control \citep{sak2015fast,chen2015deepdriving,jalab2015human,corradini2015robust}. 
Although most of deep neural networks are implemented using GPUs (Graphics Processing Units) in these days, their implementation in hardware can give many benefits in terms of power consumption and system size \citep{ovtcharov2015accelerating}. 
FPGA based implementation examples of CNN show more than 10 times advantage in power consumption \citep{ovtcharov2015accelerating}. 

Neural network algorithms employ many multiply and add (MAC) operations that mimic the operations of biological neurons. This suggests that reconfigurable hardware arrays that contain quite homogeneous hardware blocks, such as MAC units, can give very efficient solution to real-time neural network system design.  
Early studies on word-length determination of neural networks reported the needed precision of at least 8 bits \citep{holt1991back}. 
Our recent works show that the precision required for implementing FFDNN, CNN or RNN needs not be very high, especially when the quantized networks are trained again to learn the effects of lowered precision. 
In the fixed-point optimization examples shown in \citet{hwang2014fixed,anwar2015fixed,shin2015fixed}, neural networks with ternary weights showed quite good performance which was close to that of floating-point arithmetic.   

In this work, we try to know if retraining can recover the performance of FFDNN and CNN under quantization with only ternary (+1, 0, -1) levels or 3 bits (+3, +2, +1, 0, -1, -2, -3) for weight representation. Note that bias values are not quantized. 
For this study, the network complexity is changed to analyze their effects on the performance gap between floating-point and retrained low-precision fixed-point deep neural networks. 

We conduct our experiments with a feed-forward deep neural network (FFDNN) for phoneme recognition and a convolutional neural network (CNN) for image classification.
To control the network size, not only the number of units in each layer but also the number of hidden layers are varied in the FFDNN.
For the CNN, the number of feature maps for each layer and the number of layers are both changed.
The FFDNN uses the TIMIT corpus and the CNN employs the CIFAR-10 dataset.     
We also propose a metric called \emph{effective compression ratio (ECR)} for comparing extremely quantized bigger networks with moderately quantized or floating-point networks with the smaller size.
This analysis intends to find an insight to the knowledge representation capability of highly quantized networks, and also provides a guideline to network size and word-length determination for efficient hardware implementation of DNNs.  
%\begin{figure}[h]
%\begin{center}
%%\includegraphics{CNN}
%%\framebox[5.51in]{\includegraphics{CNN}}
%%\fbox{\rule[-.5cm]{0cm}{4cm}\includegraphics{CNN} \rule[-.5cm]{4cm}{0cm}}
%\includegraphics[width=0.8\linewidth]{CNN}
%%\includegraphics{CNN}
%\end{center}
%\caption{Typical CNN algorithm with N convolution layers and M fully-connected layers.}
%\label{fig_cnn}
%\end{figure}
%A typical CNN algorithm is shown in \figurename~\ref{fig_cnn}. Most CNN work in image recognition has the front network layers be convolutional, while the back network layers are fully connected. Where the front stages consist of convolution and pooling operations that extract local features. In most cases, several convolution filters are employed to yield multiple features. In other words, the total number of the convolution weights are determined by the number of feature maps. We may need to reduce the number of filters to minimize the hardware complexity at the cost of performance degradation. Since the performance of the network is affected by the size of the CNN (i.e., number of the feature maps), 

\section{Related Work}
\label{sec_related}

Fixed-point implementation of signal processing algorithms has long been of interest for VLSI based design of multimedia and communication systems.
Some of early works used statistical modeling of quantization noise for application to linear digital filters. 
The simulation-based word-length optimization method utilized simulation tools to evaluate the fixed-point performance of a system, by which non-linear algorithms can be optimized \citep{sung1995simulation}.
Ternary (+1, 0, -1) coefficients based digital filters were used to eliminate multiplications at the cost of higher quantization noise. 
The implementation of adaptive filters with ternary weights were developed, but it demanded oversampling to remove the quantization effects \citep{hussain2007short}.  

Fixed-point neural network design also has been studied with the same purpose of reducing the hardware implementation cost \citep{moerland1997neural}. 
In \citet{holt1991back}, back propagation simulation with 16-bit integer arithmetic was conducted for several problems, such as NetTalk, Parity, Protein and so on. 
This work conducted the experiments while changing the number of hidden units, which was, however, relatively small numbers.
The integer simulations showed quite good results for NetTalk and Parity, but not for Protein benchmarks.
With direct quantization of trained weights, this work also confirmed satisfactory operation of neural networks with 8-bit precision.    
An implementation with ternary weights were reported for neural network design with optical fiber networks \citep{fiesler1990weight}.
In this ternary network design, the authors employed retraining after direct quantization to improve the performance of a shallow network.

Recently, fixed-point design of DNNs is revisited, and FFDNN and CNN with ternary weights show quite good performances that are very close to the floating-point results.
The ternary weight based FFDNN and CNN are used for VLSI and FPGA based implementations, by which the algorithms can operate with only on-chip memory consuming very low power \citep{kim2014x1000}.
Binary weight based deep neural network design is also studied \citep{courbariaux2015binaryconnect}. 
Pruned floating-point weights are also utilized for efficient GPU based implementations, where small valued weights are forced to zero to reduce the number of arithmetic operations and the memory space for weight storage \citep{yu2012exploiting,han2015deep}.
A network restructuring technique using singular value decomposition technique is also studied \citep{xue2013restructuring,rigamonti2013learning}.

\section{Fixed-point FFDNN and CNN Design}
\label{sec_fixedpoint}

This section explains the design of FFDNN and CNN with varying network complexity and, also, the fixed-point optimization procedure.  

\subsection{FFDNN and CNN Design}

\begin{figure}[t]
\begin{center}
%\includegraphics{CNN}
%\framebox[5.51in]{\includegraphics{CNN}}
%\fbox{\rule[-.5cm]{0cm}{4cm}\includegraphics{CNN} \rule[-.5cm]{4cm}{0cm}}
\includegraphics[width=0.6\linewidth]{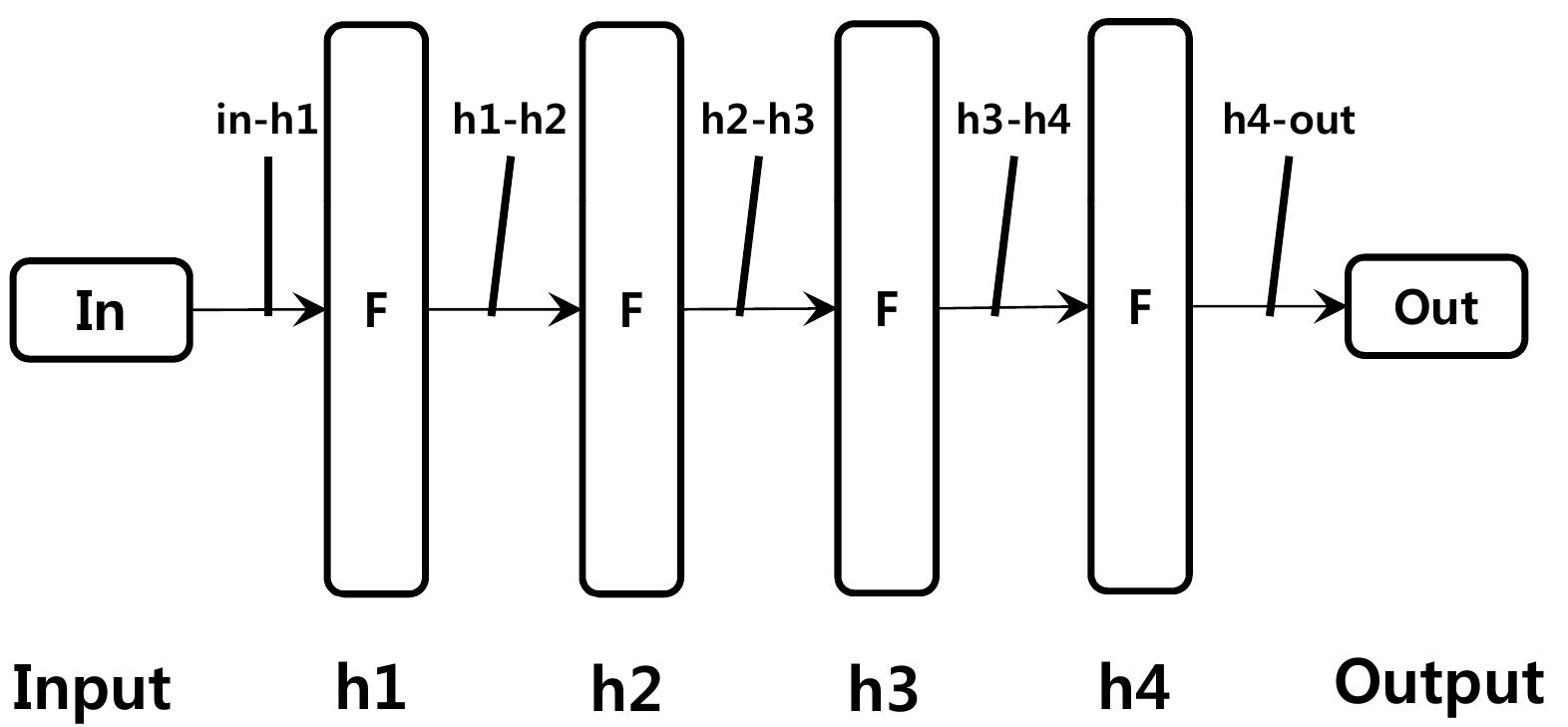}
\end{center}
\caption{Feed-forward deep neural network with 4 hidden layers.}
\label{fig_dnn}
\end{figure}

A feedforward deep neural network with multiple hidden layers are depicted in \figurename~\ref{fig_dnn}. 
Each layer $k$ has a signal vector $y_{k}$, which is propagated to the next layer by multiplying the weight matrix $W_{k+1}$, adding biases $b_{k+1}$, and applying the activation function $\phi_{k+1}(\cdot)$ as follows:
\begin{align}
	y_{k+1} = \phi_{k+1}\bigl(W_{k+1}y_{k}+b_{k+1}\bigr). \label{eq:matrixmultiple}
\end{align}
One of the most popular activation functions is the rectified linear unit defined as 
\begin{align}
	Relu(x) = max(0,x)   \label{eq:reru}.
\end{align}

In this work, an FFDNN for phoneme recognition is used. The reference DNN has four hidden layers. Each of the hidden layers has $N_{h}$ units; the value of $N_{h}$ is changed to control the complexity of the network. 
We conduct experiments with the $N_{h}$ size of 32, 64, 128, 256, 512, and 1024. The number of hidden layers is also reduced. 
The input layer of the network has 1,353 units to accept 11 frames of a Fourier-transform-based filter-bank with 40 coefficients ($+$energy) distributed on a mel-scale, together with their first and second temporal derivatives. The output layer consists of 61 softmax units which correspond to 61 target phoneme labels. Phoneme recognition experiments were performed on the TIMIT corpus. The standard 462 speaker set with all SA records removed was used for training, and a separate development set of 50 speaker was used for early stopping. Results are reported for the 24-speaker core test set. The network was trained using a backpropagation algorithm with 128 mini-batch size. Initial learning rate was $10^{-5}$ and it was decreased until $10^{-7}$ during the training. Momentum was 0.9 and RMSProp was adopted for weights update \citep{tieleman2012lecture}. 
The dropout technique was employed with 0.2 dropout rate in each layer. 

\begin{figure}[t]
\begin{center}
%\includegraphics{CNN}
%\framebox[5.51in]{\includegraphics{CNN}}
%\fbox{\rule[-.5cm]{0cm}{4cm}\includegraphics{CNN} \rule[-.5cm]{4cm}{0cm}}
\includegraphics[width=0.8\linewidth]{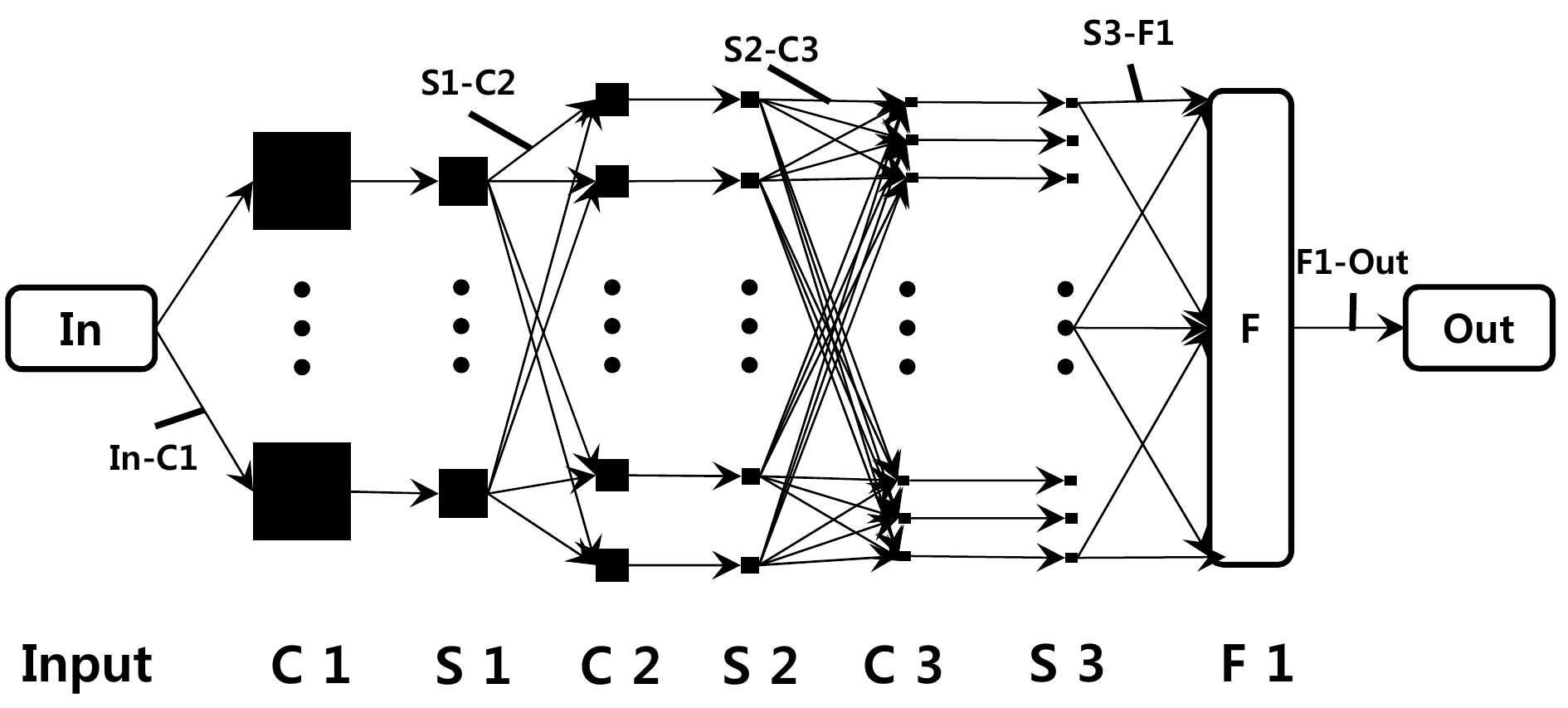}
\end{center}
\caption{CNN structure with 3 convolution layers and 1 fully-connected layers.}
\label{fig_cnn}
\end{figure}

The CNN used is for CIFAR-10 dataset. 
It contains a training set of 50,000 and a test set of 10,000 32$\times$32 RGB color images representing airplanes, automobiles, birds, cats, deers, dogs, frogs, horses, ships and trucks. We divided the training set to 40,000 images for training and 10,000 images for validation. This CNN has 3 convolution and pooling layers and a fully connected hidden layer with 64 units, and the output has 10 softmax units as shown in \figurename~\ref{fig_cnn}.
We control the number of feature maps in each convolution layer. The reference size has 32-32-64 feature maps with 5 by 5 kernel size as used in~\citet{krizhevskey2014cuda}. We did not perform any preprocessing and data augmentation such as ZCA whitening and global contrast normalization. 
To know the effects of network size variation, the number of feature maps is reduced or increased. 
The configurations of the feature maps used for the experiments are 8-8-16, 16-16-32, 32-32-64, 64-64-128, 96-96-192, and 128-128-256. 
The number of feature map layers is also changed, resulting in 32-32-64, 32-64, and 64 map configurations. 
Note that the fully connected layer in the CNN is not changed. 
The network was trained using a backpropagation algorithm with 128 mini-batch size. Initial learning rate was 0.001 and it was decreased to $10^{-8}$ during the training procedure. Momentum was 0.8 and RMSProp was applied for weights update. 

\subsection{Fixed-point optimization of DNNs}

Reducing the word-length of weights brings several advantages in hardware based implementation of neural networks.
First, it lowers the arithmetic precision, and thereby reduces the number of gates needed for multipliers.  
Second, the size of memory for storing weights is minimized, which would be a big advantage when keeping them on a chip, instead of external DRAM or NAND flash memory.
Note that FFDNNs and recurrent neural networks demand a very large number of weights.
Third, the reduced arithmetic precision or minimization of off-chip memory accesses leads to low power consumption.
However, we need to concern the quantization effects that degrade the system performance.

Direct quantization converts a floating-point value to the closest integer number, which is conventionally used in signal processing system design. 
However, direct quantization usually demands more than 8 bits, and does not show good performance when the number of bits is small. 
In fixed-point deep neural network design, retraining of quantized weights shows quite good performance.

The fixed-point DNN algorithm design consists of three steps: floating-point training, direct quantization, and retraining of weights.
The floating-point training procedure can be any of the state of the art techniques, which may include unsupervised learning and dropout.
Note that fixed-point optimization needs to be based on the best performing floating-point weights.  
Thus, the floating-point weight optimization may need to be conducted several times with different initializations, and this step consumes the most of the time.
After the floating-point training, direct quantization is followed.

For direct quantization, uniform quantization function is employed and the function $Q(\cdot)$ is defined as follows :
\begin{align}
	Q(w) = sgn(w) \cdot \Delta  \cdot min\biggl(\left \lfloor{\dfrac{\lvert(w)\rvert}\Delta +0.5}\right \rfloor , \dfrac{M-1}2\biggr) \label{eq:uniform_quant}
\end{align}  
where $sgn(\cdot)$ is a sign function, $\Delta$ is a quantization step size, and $M$ represents the number of quantization levels. 
Note that $M$ needs to be an odd number since the weight values can be positive or negative. 
When $M$ is 7, the weights are represented by -3$\cdot\Delta$, -2$\cdot\Delta$,  -1$\cdot\Delta$, 0, +1$\cdot\Delta$, +2$\cdot\Delta$, +3$\cdot\Delta$,which can be represented in 3 bits. 
\\
The quantization step size $\Delta$ is determined to minimize the L2 error, $E$, depicted as follows. 
\begin{align}
	E = \dfrac1 2\sum\limits_{i=1}^N\bigl(Q(w_{i})-w_{i} \bigr)^2   \label{eq:error}
\end{align}
where $N$ is the number of weights in each weight group, $w_{i}$ is the $i$-th weight value represented in floating-point. 
This process needs some iterations, but does not take much time.  
%Bias values are not quantized.

\indent For network retraining, we maintain both floating-point and quantized weights because the amount of weight updates in each training step is much smaller than the quantization step size $\Delta$. 
The forward and backward propagation is conducted using quantized weights, but the weight update is applied to the floating-point weights and newly quantized values are generated at each iteration.
This retraining procedure usually converges quickly and does not take much time when compared to the floating-point training.

\section{Analysis of quantization effects}
\label{sec:experiments}

\subsection{Direct quantization}
The performance of the FFDNN and the CNN with directly quantized weights is analyzed while varying the number of units in each layer or the number of feature maps, respectively.
In this analysis, the quantization is performed on each weight group, which is illustrated in \figurename~\ref{fig_dnn} and \figurename~\ref{fig_cnn}, to know the sensitivity of word-length reduction. 
In this sub-section, we try to analyze the effects of direct quantization. 

The quantized weight can be represented as follows, 
\begin{align}
	w_{i}^q = w_{i} + w_{i}^d
\end{align}

where $w_{i}^d$ is the distortion of each weight due to quantization.  
In the direct quantization, we can assume that the distortion $w_{i}^d$ is not dependent each other.

\begin{figure}[h]
\centering
\subfloat[][]{\includegraphics[width=0.3\linewidth]{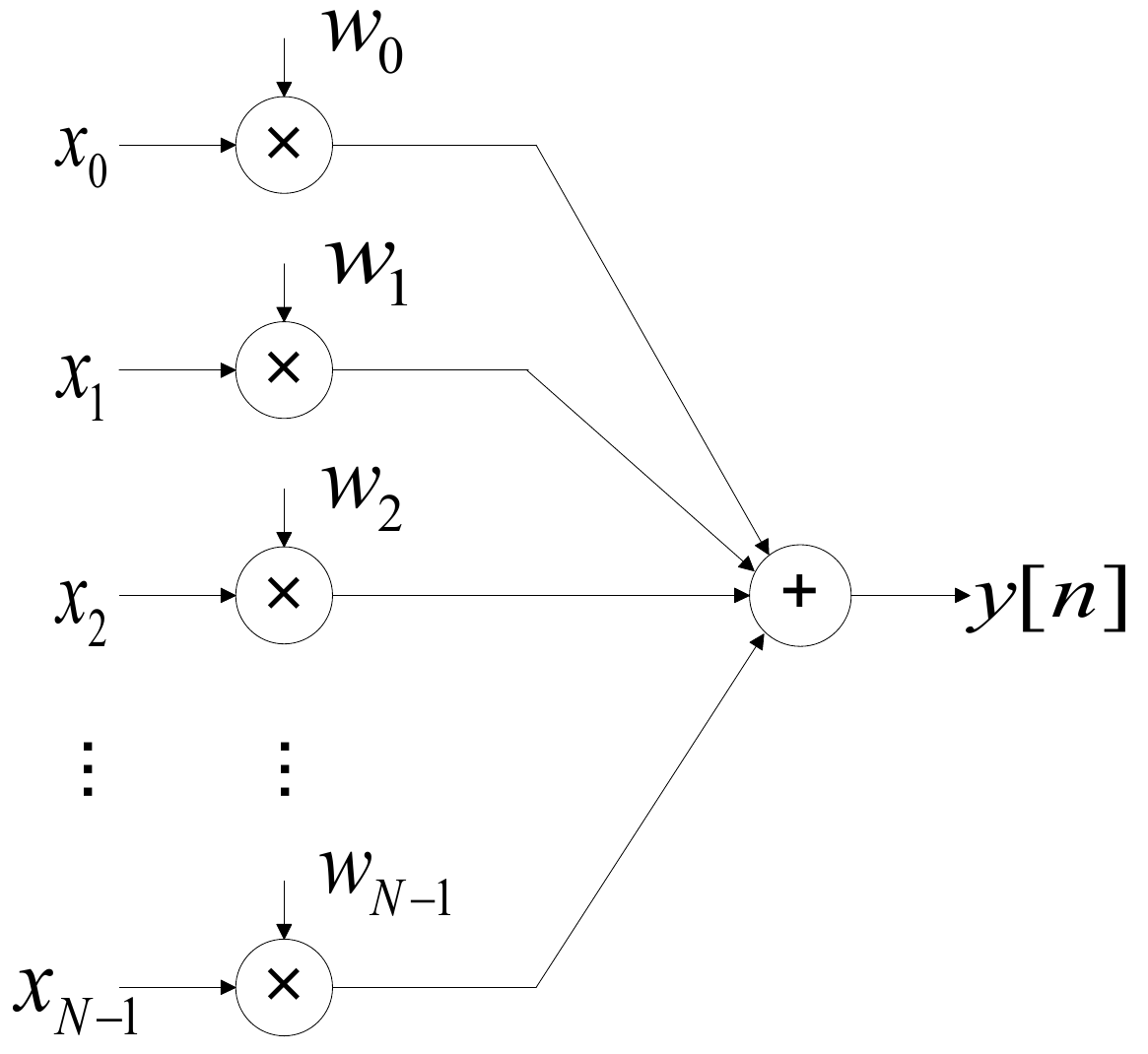}\label{fig_adder_a}}
\quad\quad\quad
\subfloat[][]{\includegraphics[width=0.3\linewidth]{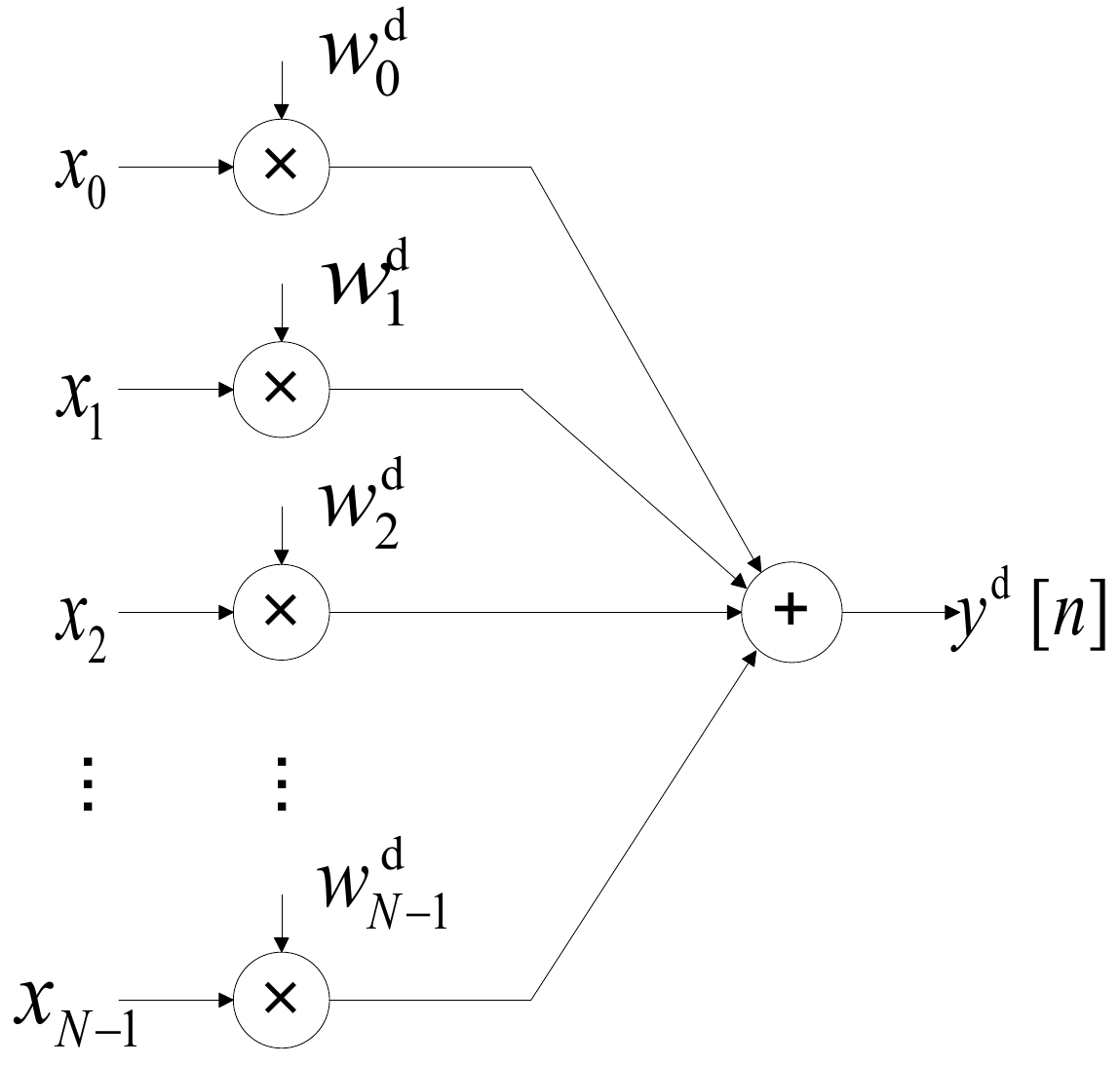}\label{fig_adder_b}}
\caption{Computation model for a unit in the hidden layer $j$ ((a): floating-point, (b): distortion).}
\label{fig_adder}
\end{figure}

\begin{figure}[h]
\centering
\subfloat[][]{\includegraphics[width=0.495\linewidth]{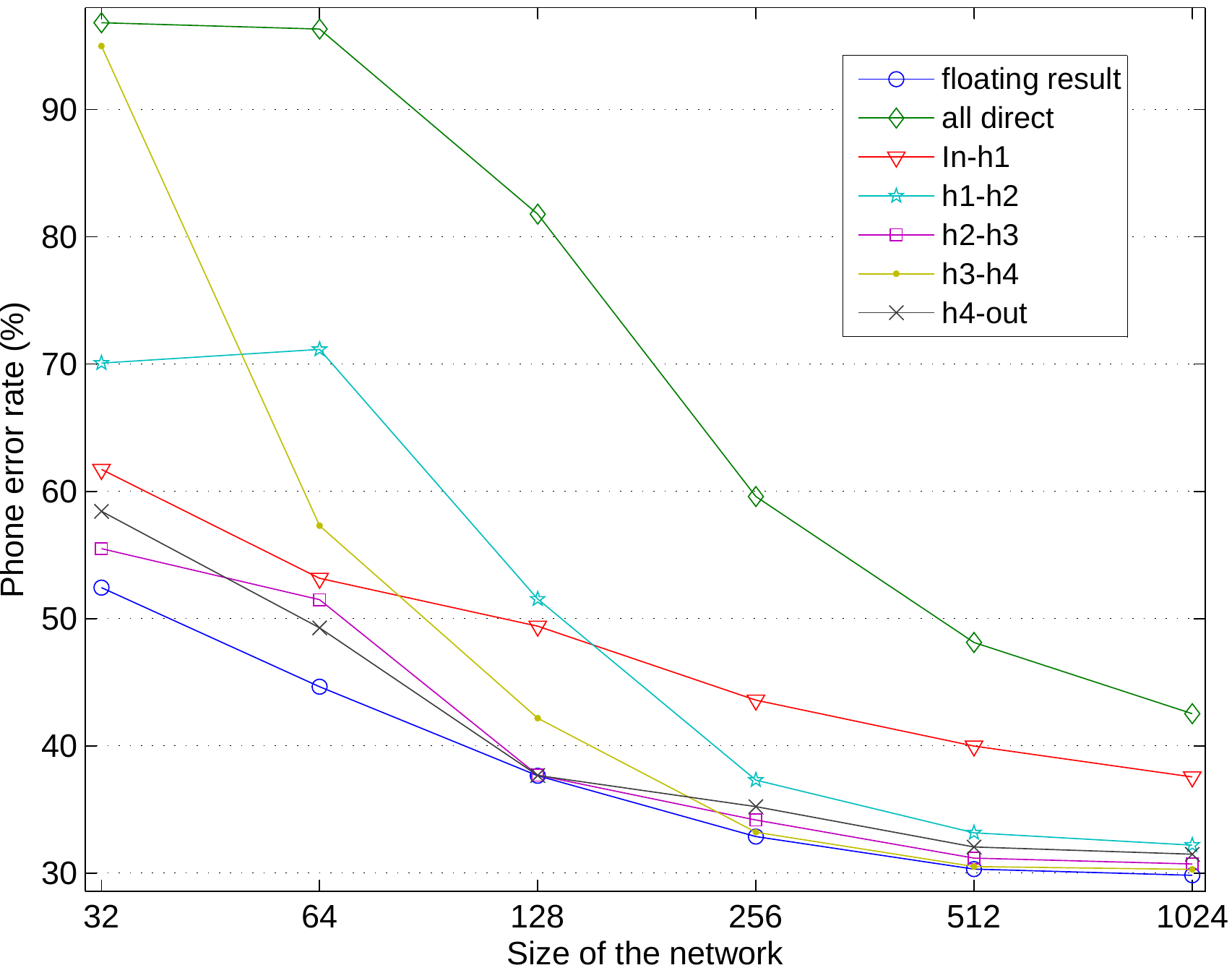}\label{direct_sensitivity_phone}}
\subfloat[][]{\includegraphics[width=0.505\linewidth]{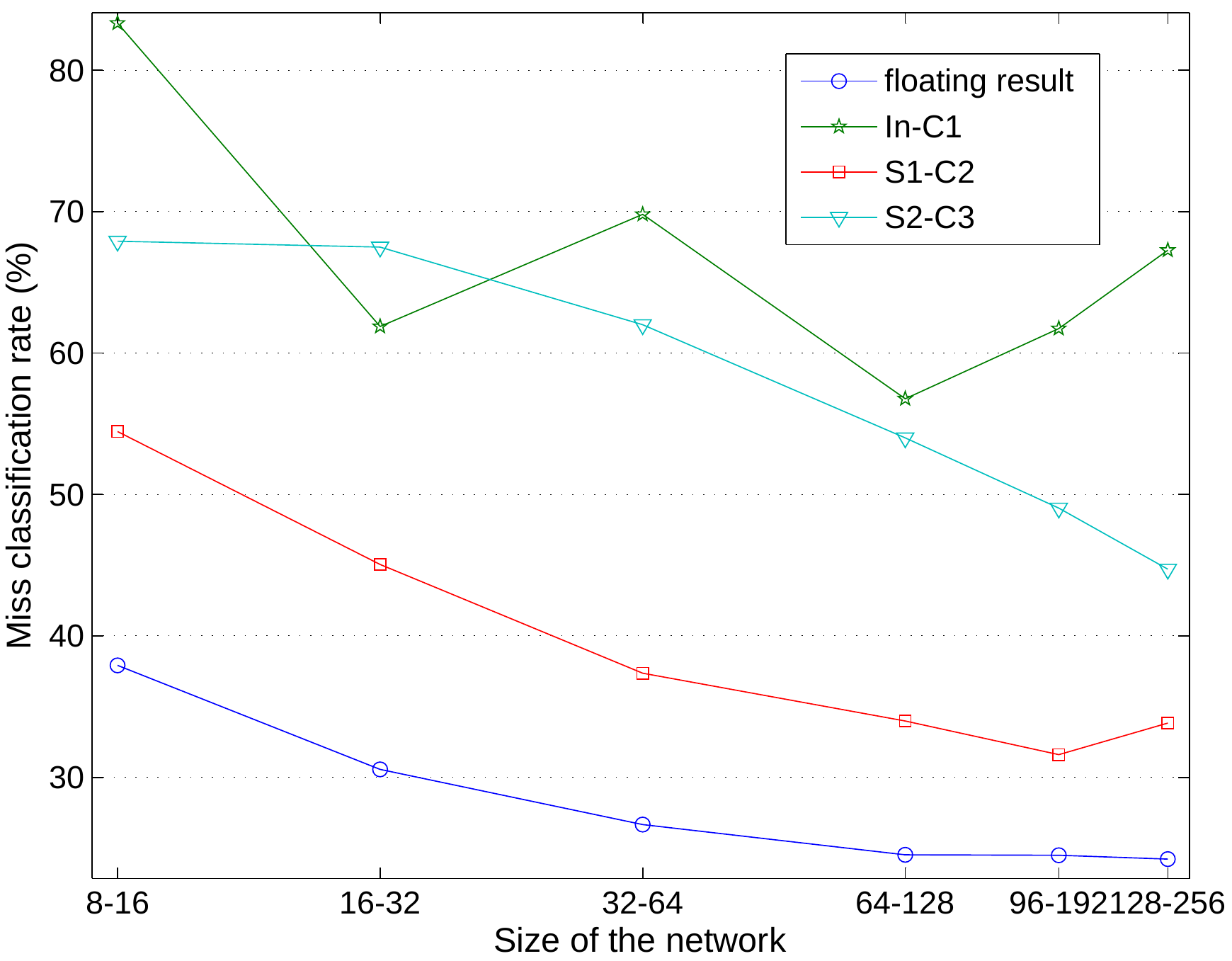}\label{direct_sensitivity_cnn2bit}}
\caption{Sensitivity analysis of direct quantization ((a): FFDNN, (b): CNN). In the figure (b), x-axis label `8-16' represents the number of feature map is `8-8-16'.}
\label{fig_direct_sensitivity}
\end{figure}

Consider a computation procedure for a unit in a hidden layer, the signal from the previous layer is summed up after multiplication with the weights as illustrated in \figurename~\ref{fig_adder_a}.
We can also assemble a model for distortion, which is shown in \figurename~\ref{fig_adder_b}.
In the distortion model, since $w_{i}^d$ is independent each other, we can assume that the effects of the summed distortion is reduced according to the random process theory.
This analysis means that the quantization effects are reduced when the number of units in the anterior layer increases, but slowly.

\begin{figure}[h]
\centering
\subfloat[][]{\includegraphics[width=0.5\linewidth]{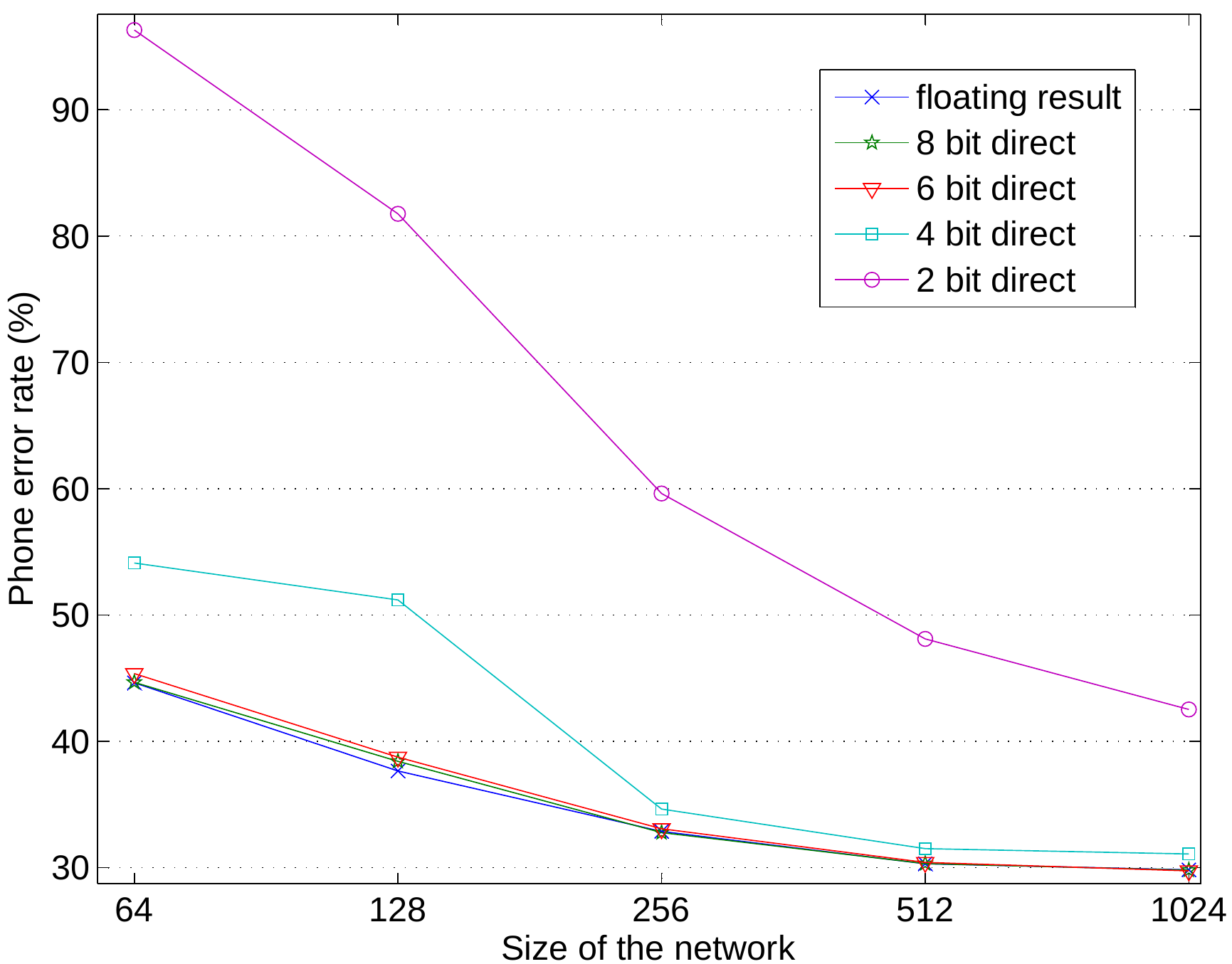}\label{fig_direct_only_a}}
\subfloat[][]{\includegraphics[width=0.5\linewidth]{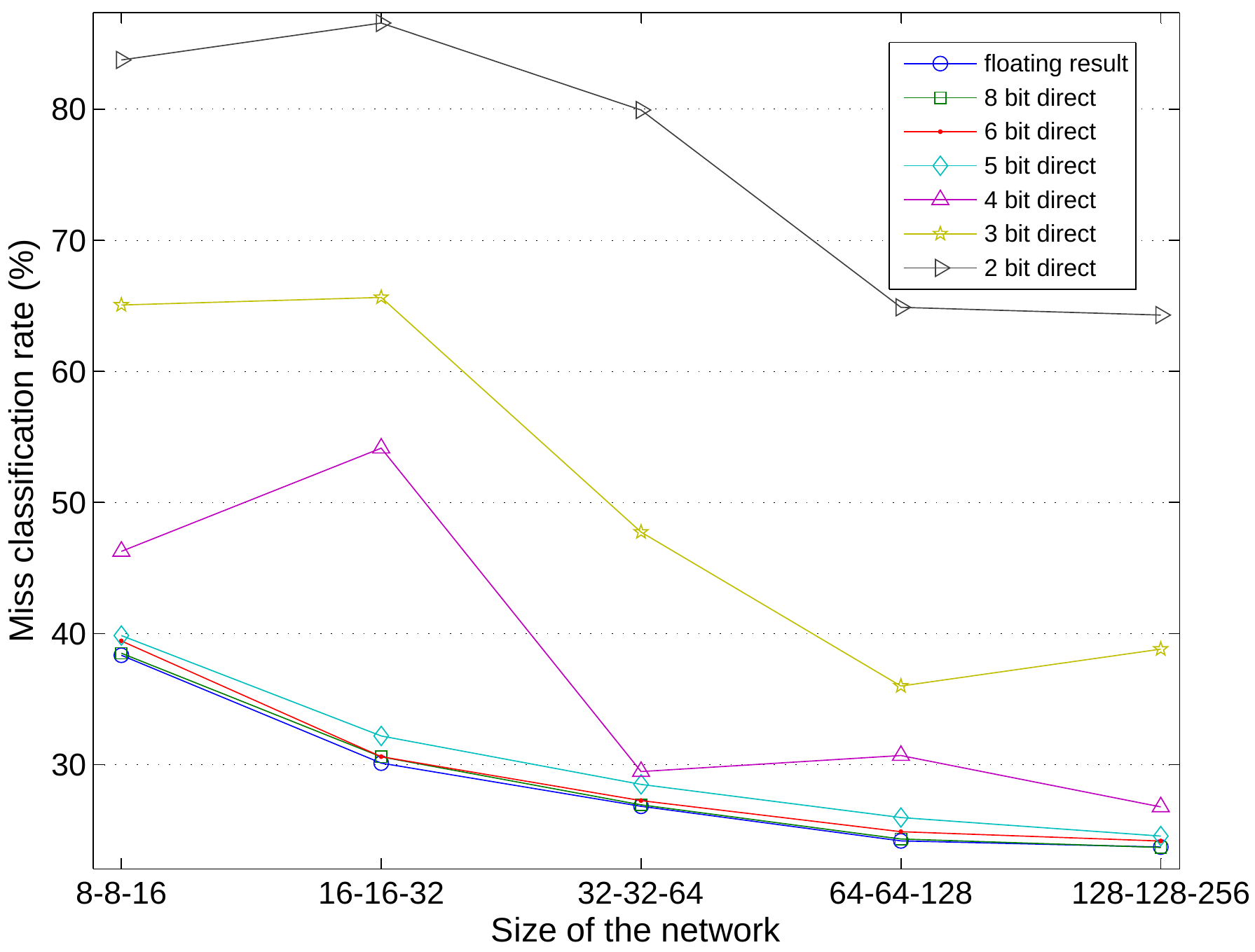}\label{fig_direct_only_b}}
\caption{Performance of direct quantization with multiple precision ((a): FFDNN, (b): CNN).}
\label{fig_direct_only}
\end{figure}

\figurename~\ref{direct_sensitivity_phone} illustrates the performance of the FFDNN with floating-point arithmetic, 2-bit direct quantization of all the weights, and 2-bit direct quantization only on the weight group `In-h1', `h1-h2', and `h4-out'.
Consider the quantization performance of the `In-h1' layer, the phone-error rate is higher than the floating-point result with an almost constant amount, about 10\%. 
Note that the number of input to the `In-h1' layer is fixed, 1353, regardless of the hidden unit size. 
Thus, the amount of distortion delivered to each unit of the hidden layer 1 can be considered unchanged.
\figurename~\ref{direct_sensitivity_phone} also shows the quantization performance on `h1-h2' and `h4-out' layers, which informs the trend of reduced gap to the floating-point performance as the network size increases.
This can be explained by the sum of increased number of independent distortions when the network size grows. 
The performance of all 2-bit quantization also shows the similar trend of reduced gap to the floating-point performance. 
But, apparently, the performance of 2-bit directly quantized networks is not satisfactory.

In \figurename~\ref{direct_sensitivity_cnn2bit}, a similar analysis is conducted to the CNN with direct quantization when the number of feature maps increases or decreases. 
In the CNN, the number of input to each output is determined by the number of input feature maps and the kernel size. 
For example, at the first layer C1, the number of input signal for computing one output is only 75 (=3$\times$25) regardless of the network size, where the input map size is always 3 and the kernel size is 25. 
However, at the second layer C2, the number of input feature maps increases as the network size grows.   
When the feature map of 32-32-64 is considered, the number of input for the C2 layer grows to 800 (=32$\times$25).
Thus, we can expect a reduced distortion as the number of feature maps increases. 

\figurename~\ref{fig_direct_only_a} shows the performance of direct quantization with 2, 4, 6, and 8-bit precision when the network complexity varies.  In the FFDNN, 6 bit direct quantization seems enough when the network size is larger than 128. But, small FFDNNs demand 8 bits for near floating-point performance. The CNN in \figurename~\ref{fig_direct_only_b} also shows the similar trend.  The direct quantization requires about 6 bits when the feature map configuration is 16-16-32 or larger.
  
\subsection{Effects of retraining on quantized networks}

\begin{figure}[h]
\centering
\subfloat[][]{\includegraphics[width=0.5\linewidth]{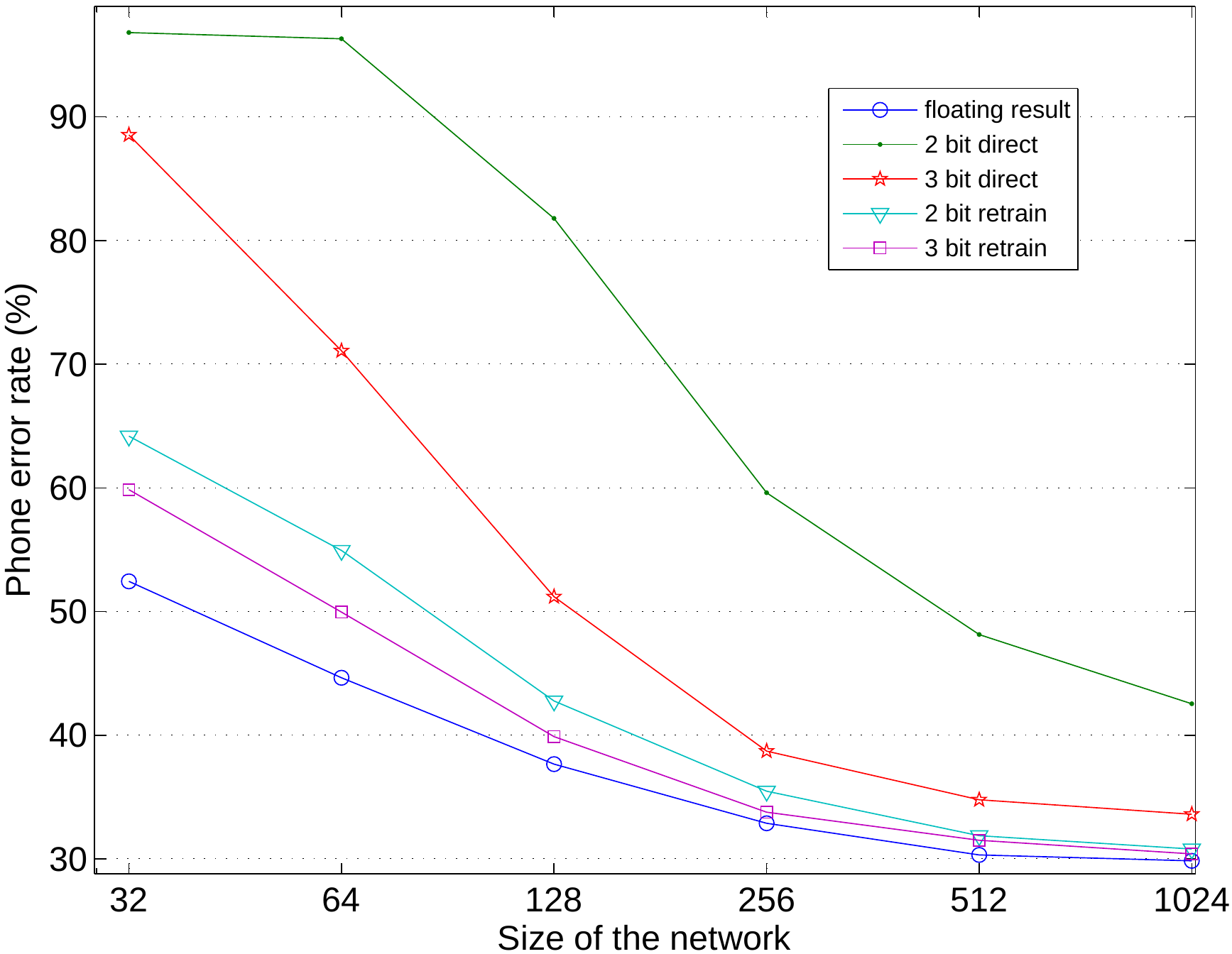}\label{fig_dnn_full_quant}}
\subfloat[][]{\includegraphics[width=0.5\linewidth]{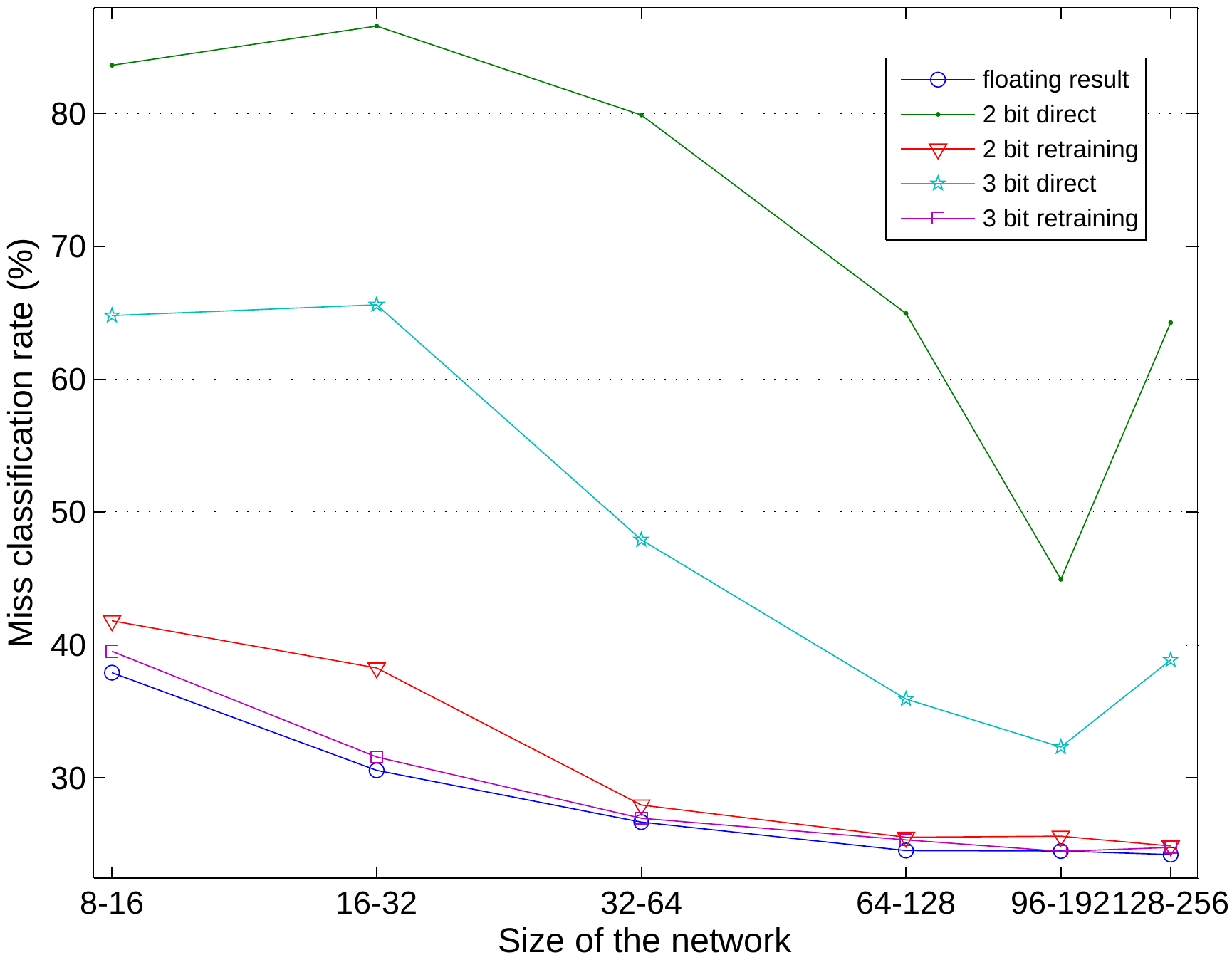}\label{fig_cnn_full_quant}}
\caption{Comparison of retrain-based and direct quantization for DNN (a) and CNN (b). All the weights are quantized with ternary and 7-level weights. In the figure (b), x-axis label '8-16' represents the number of feature map is '8-8-16'.}
\label{fig_full_quant}
\end{figure}

Retraining is conducted on the directly quantized networks using the same data for floating-point training.
The fixed-point performance of the FFDNN is shown in \figurename~\ref{fig_dnn_full_quant} when the number of hidden units in each layer varies. 
The performance of direct 2 bits (ternary levels), direct 3 bits (7-levels), retrain-based 2 bits, and retrain-based 3 bits are compared with the floating-point simulation.
We can find that the performance gap between the floating-point and the retrain-based fixed-point networks converges very fast as the network size grows. 
Although the performance gap between the direct and the floating-point networks also converges, the rate of convergence is significantly different. 
In this figure, the performance of the floating-point network almost saturates when the network size is about 1024. 
Note that the TIMIT corpus that is used for training has only 3 hours of data. 
Thus, the network with 1024 hidden units can be considered in the `training-data limited region'. 
Here, the gap between the floating-point and fixed-point networks almost vanishes when the network is in the `training-data limited region'.
However, when the network size is limited, such as 32, 64, 128, or 256, there is some performance gap between the floating-point and highly quantized networks even if retraining on the quantized networks is performed. 
%\begin{figure}[h]
%\begin{minipage}[t]{0.5\linewidth}
%\includegraphics[width=\linewidth]{real_2bit3bit_direct_retraining}
%\label{f1}
%\end{minipage}
%\hfill
%\begin{minipage}[t]{0.5\linewidth}
%\includegraphics[width=\linewidth]{real_2bit3bit_direct_retraining_cnn}
%\label{f2}
%\end{minipage}
%\caption{Quantization sensitivity analysis of FFDNN (left) and CNN (right) with varying network size. Ternary weights generated with retraining are used for each named group, while other groups are not quantized.}
%\label{fig_dnn_sensitivity}
%\end{figure}

%\begin{figure}[h]
%\begin{center}
%\includegraphics[width=0.8\linewidth]{real_2bit3bit_direct_retraining}
%\end{center}
%\caption{Comparison of retrain-based and direct quantization. All the weights are quantized with ternary and 7-level weights.}
%\label{fig_dnn_full_quant}
%\end{figure}

The similar experiments are conducted for the CNN with varying feature map sizes, and the results are shown in \figurename~\ref{fig_cnn_full_quant}.
The configuration of the feature maps used for the experiments are 8-8-16, 16-16-32, 32-32-64, 64-64-128, 96-96-192, and 128-128-256. 
The size of the fully connected layer is not changed.
In this figure, the floating-point and the fixed-point performances with retraining also converge very fast as the number of feature maps increases. 
The floating-point performance saturates when the feature map size is 128-128-256, and the gap is less than 1\% when comparing the floating-point and the retrain-based 2-bit networks.
However, also, there is some performance gap when the number of feature maps is reduced.  
This suggests that a fairly high performance feature extraction can be designed even using very low-precision weights if the number of feature maps can be increased.

%\begin{figure}[h]
%\begin{center}
%\includegraphics[width=0.8\linewidth]{real_2bit3bit_direct_retraining_cnn}
%\end{center}
%\caption{Comparison of retrain-based and direct quantization for CNN. All the weights are quantized with ternary and 7-level weights.}
%\label{fig_cnn_quant}
%\end{figure}

\subsection{Fixed-point performances when varying the depth}
It is well known that increasing the depth usually results in positive effects on the performance of a DNN \citep{yu2012more}.
The network complexity of a DNN is changed by increasing or reducing the number of hidden layers or feature map levels. 
The result of fixed-point and floating-point performances when varying the number of hidden layers for the FFDNN is summarized in~\tablename~\ref{table_dnn2}.  The number of units in each hidden layer is 512.  
This table shows that both the floating-point and the fixed-point performances of the FFDNN increase when adding hidden layers from 0 to 4.
The performance gap between the floating-point and the fixed-point networks shrinks as the number of levels increases. 

\begin{table}[h]
\centering
\caption{Framewise phoneme error rate on TIMIT with respect to the depth in DNN}
\label{table_dnn2}
\begin{tabular}{ c c c c c }
\hline
\textbf{\begin{tabular}[c]{@{}c@{}}Number of layers\\ (Floating-point result)\end{tabular}}                                                         & \textbf{\# Quantization levels} & \textbf{Direct} & \textbf{Retraining}&\textbf{Difference} \\ \hline \hline

\multirow{2}{*}{\begin{tabular}[c]{@{}c@{}}1\\ (34.67\%)\end{tabular}} & 3-level              & 69.88\%         & 38.58\%    &     3.91\%    \\ 
                                                                       & 7-level              & 56.81\%         & 36.57\%       &   1.90\%    \\ \hline
\multirow{2}{*}{\begin{tabular}[c]{@{}c@{}}2\\ (31.51\%)\end{tabular}} & 3-level              & 47.74\%         & 33.89\%     &    2.38\%    \\ 
                                                                       & 7-level              & 36.99\%         & 33.04\%    &     1.53\%    \\ \hline
\multirow{2}{*}{\begin{tabular}[c]{@{}c@{}}3\\ (30.81\%)\end{tabular}} & 3-level              & 49.27\%         & 33.05\%     &  2.24\%      \\ 
                                                                       & 7-level              & 36.58\%         & 31.72\%    &      0.91\%   \\ \hline
\multirow{2}{*}{\begin{tabular}[c]{@{}c@{}}4\\ (30.31\%)\end{tabular}} & 3-level              & 48.13\%         & 31.86\%     &     1.55\%   \\ 
                                                                       & 7-level              & 34.77\%         & 31.49\%     &   1.18\%     \\ \hline
\end{tabular}
\end{table}

The network complexity of the CNN is also varied by reducing the level of feature maps as shown in~\tablename~\ref{table_cnn2}.  
As expected, the performance of both the floating-point and retrain-based low-precision networks degrades as the number of levels is reduced. 
The performance gap between them is very small with 7-level quantization for all feature map levels.   

These results for the FFDNN and the CNN with varied number of levels also show that the effects of quantization can be much reduced by retraining when the network contains some redundant complexity. 

\begin{table}[h]
\centering
\caption{Miss classification rate on CIFAR-10 with respect to the depth in CNN}
\label{table_cnn2}
\begin{tabular}{c c c c c}
\hline
\textbf{\begin{tabular}[c]{@{}c@{}c@{}}Layer\\ (Floating-point result)\end{tabular}}   & \textbf{\# Quantization levels} & \textbf{Direct}& \textbf{Retraining} &\textbf{Difference} \\ \hline \hline
\multirow{2}{*}{\begin{tabular}[c]{@{}c@{}c@{}}64\\ (34.19\%)\end{tabular}}       & 3-level              & 72.95\%         & 35.37\%    &    1.18\%    \\
                                                                              & 7-level              & 46.60\%         & 34.15\%         &   -0.04\%  \\ \hline
\multirow{2}{*}{\begin{tabular}[c]{@{}c@{}c@{}}32-64\\ (29.29\%)\end{tabular}}    & 3-level              & 55.30\%         & 29.51\%    &    0.22\%     \\ 
                                                                              & 7-level              & 39.80\%         & 29.32\%          & 0.03\%  \\ \hline
\multirow{2}{*}{\begin{tabular}[c]{@{}c@{}c@{}}32-32-64\\ (26.87\%)\end{tabular}} & 3-level              & 79.88\%         & 27.94\%    &   1.07\%     \\
                                                                              & 7-level              & 47.91\%         & 26.95\%         & 0.08\%  \\ \hline
\end{tabular}
\end{table}

\section{Effective compression ratio}

So far we have examined the effect of direct and retraining-based quantization to the final classification error rates. As the number of quantization level decreases, more memory space can be saved at the cost of sacrificing the accuracy. Therefore, there is a trade-off between the total memory space for storing weights and the final classification accuracy. In practice, investigating this trade-off is important for deciding the optimal bit-widths for representing weights and implementing the most efficient neural network hardware.

In this section, we propose a guideline for finding the optimal bit-widths in terms of the total number of bits consumed by the network weights when the desired accuracy or the network size is given. Note that we assume $2n - 1$ quantization levels are represented by $n$ bits (i.e. 2 bits are required for representing a ternary weight). For simplicity, all layers are quantized with the same number of quantization levels. However, the similar approach can be applied to the layer-wise quantization analysis.

\begin{figure}[h]
\centering
\subfloat[][]{\includegraphics[width=0.5\linewidth]{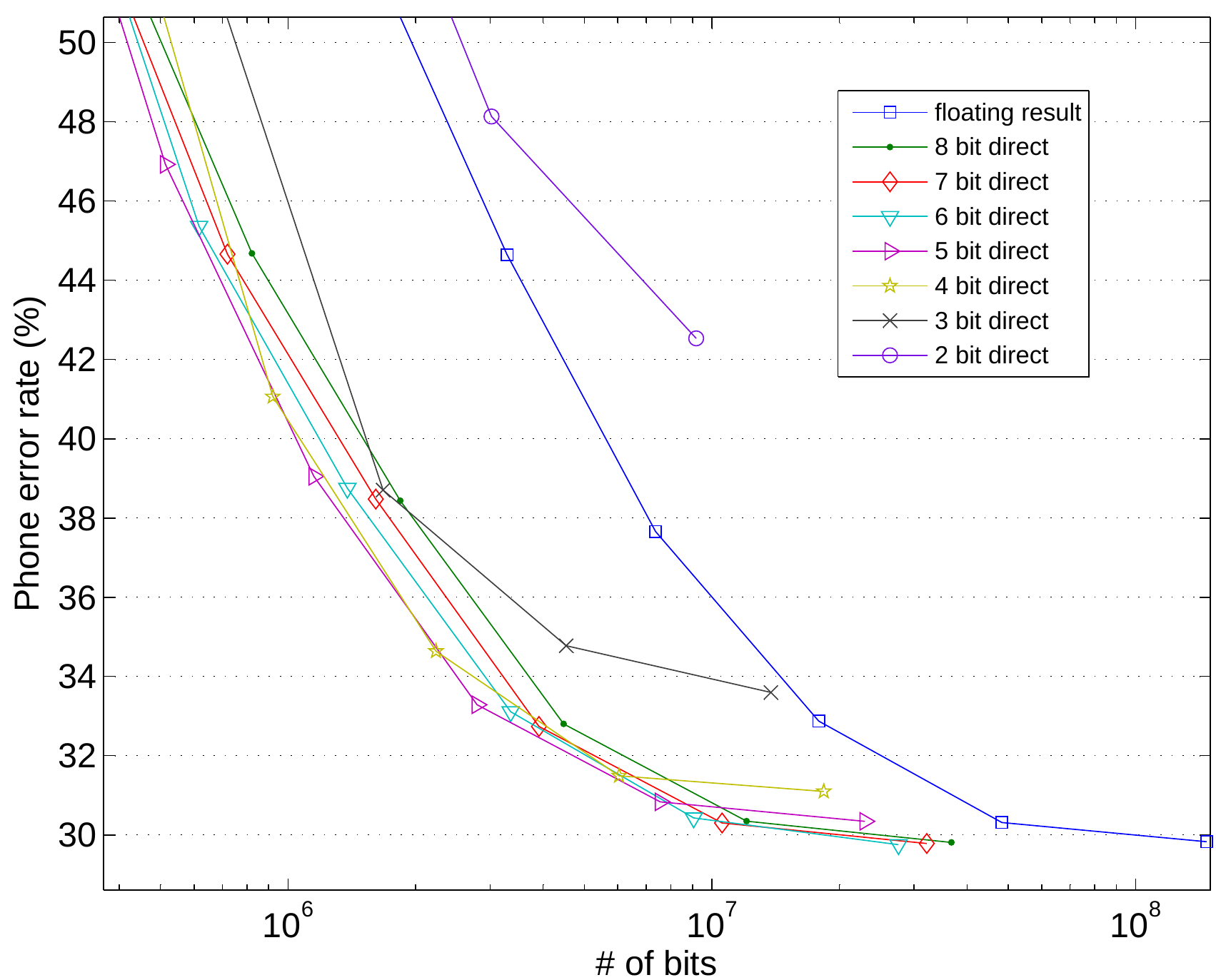}\label{fig_number_of_bits_a}}
\subfloat[][]{\includegraphics[width=0.5\linewidth]{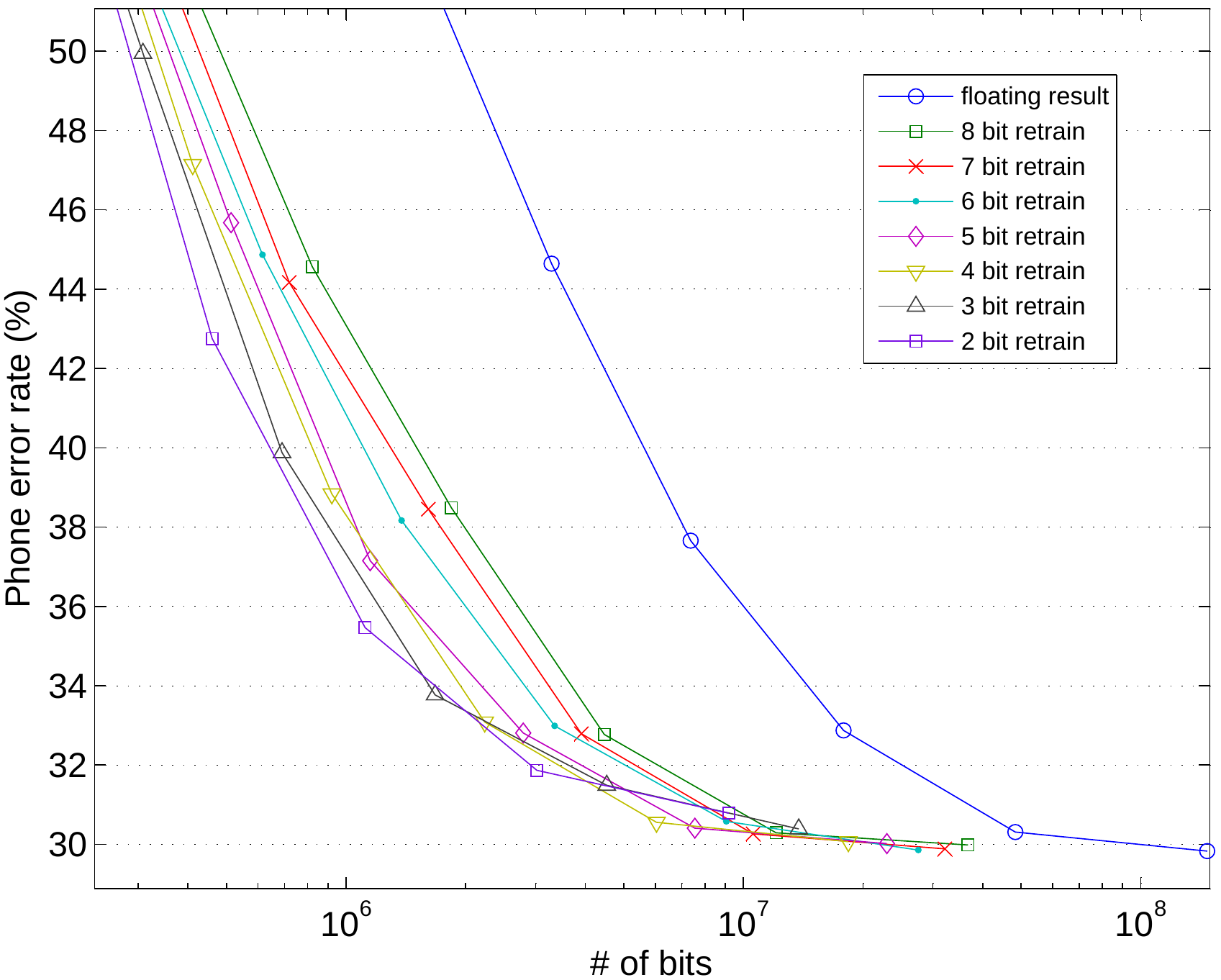}\label{fig_number_of_bits_b}}
\caption{Framewise phone error rate of phoneme recognition DNNs with respect to the total number of bits for weights with (a) direct quantization and (b) after retraining.}
\label{fig_number_of_bits}
\end{figure}

The optimal combination of the bit-width and layer size can be found when the number of total bits or the accuracy is given as shown in \figurename~\ref{fig_number_of_bits}. The figure shows the framewise phoneme error rate on TIMIT with respect to the number of total bits, while varying the layer size of DNNs with various number of quantization bits from 2 to 8 bits. The network has 4 hidden layers with the uniform sizes. With direct quantization, the optimal hardware design can be achieved with about 5 bits. On the other hand, the weight representation with only 2 bits shows the best performance after retraining.

\begin{figure}[h]
\centering
\includegraphics[width=0.5\linewidth]{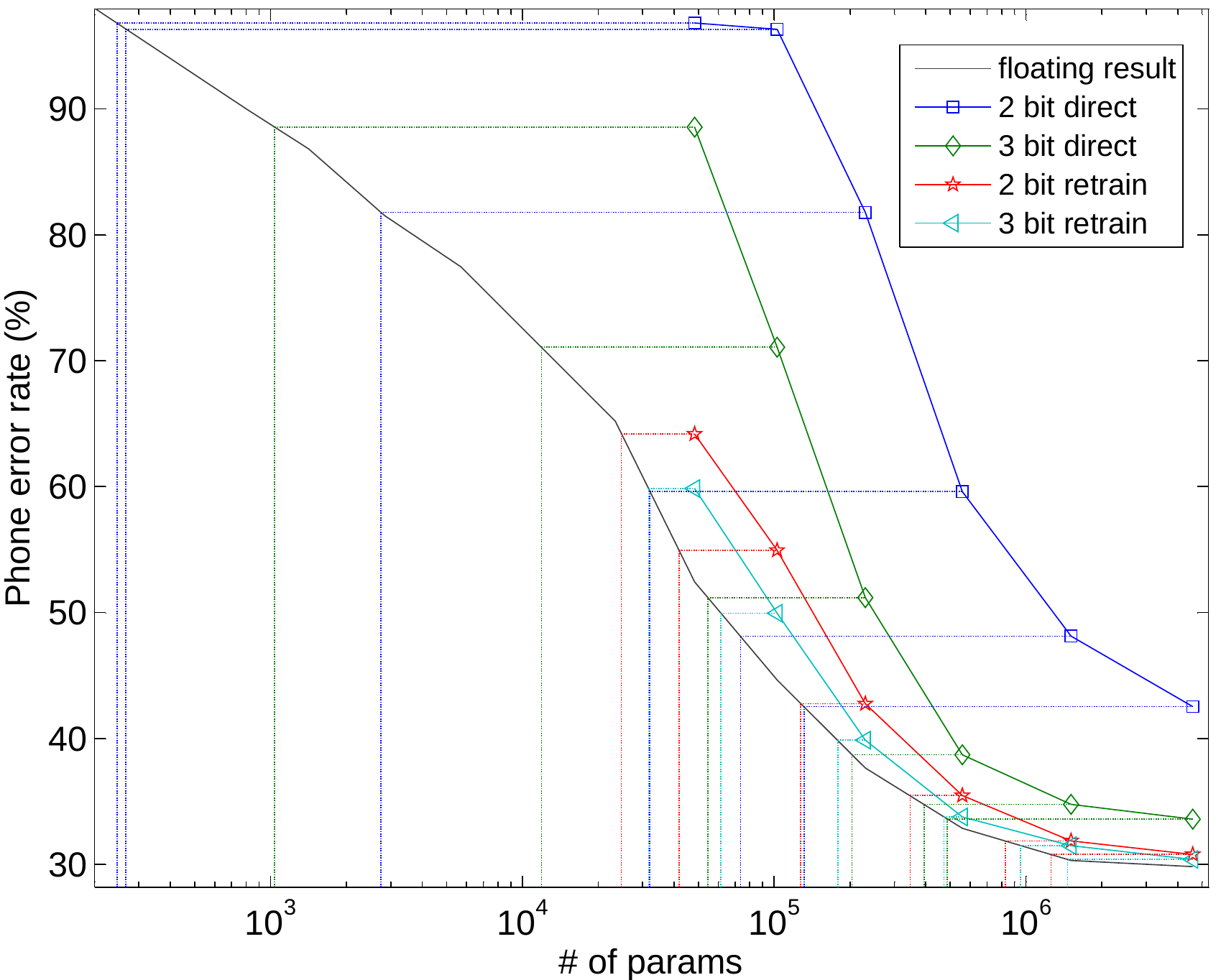}
\caption{Obtaining effective number of parameters for the uncompressed network.}
\label{fig_ecr_lines}
\end{figure}

\begin{figure}[h]
\centering
\subfloat[][]{\includegraphics[width=0.5\linewidth]{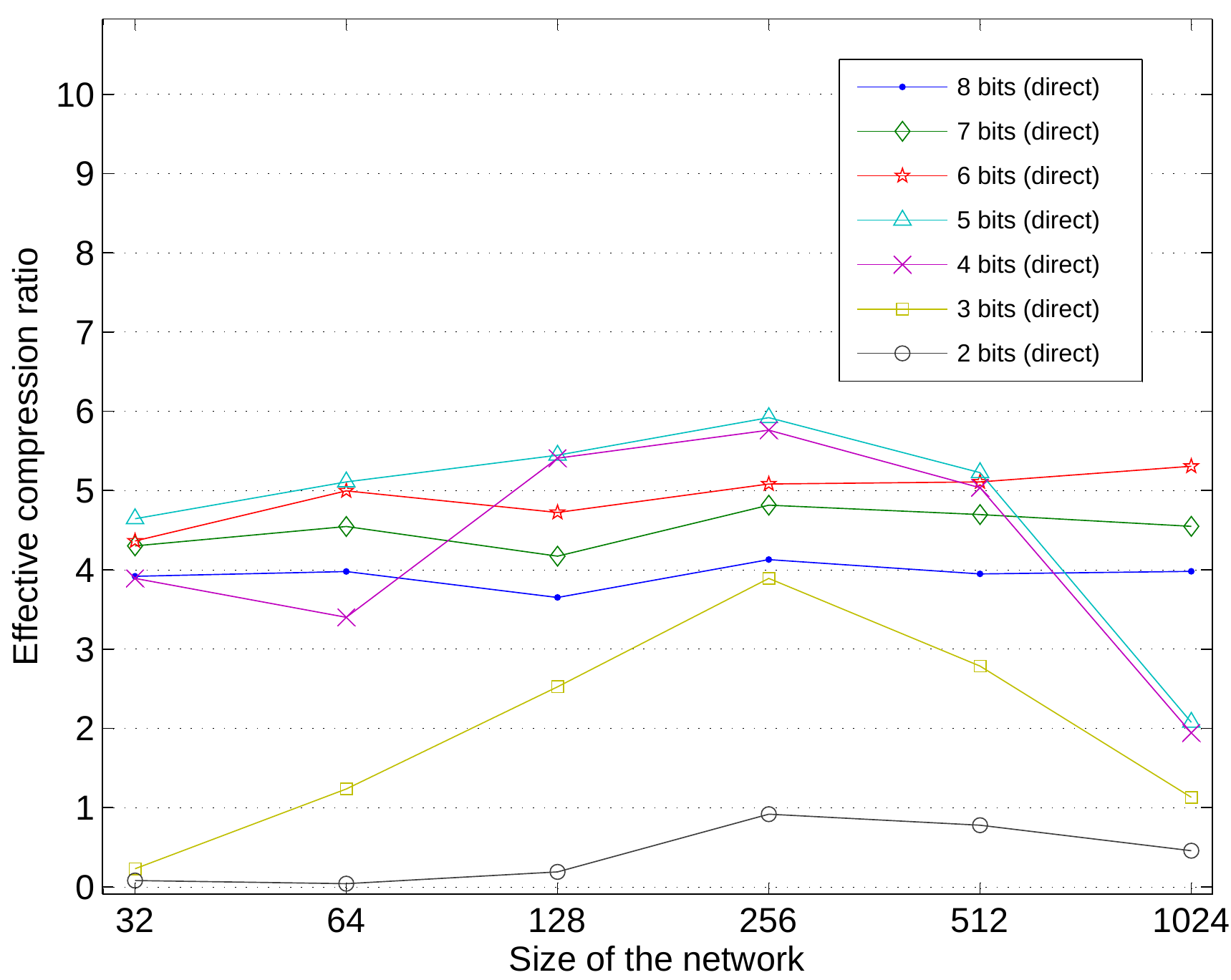}\label{fig_ecr_direct}}
\subfloat[][]{\includegraphics[width=0.5\linewidth]{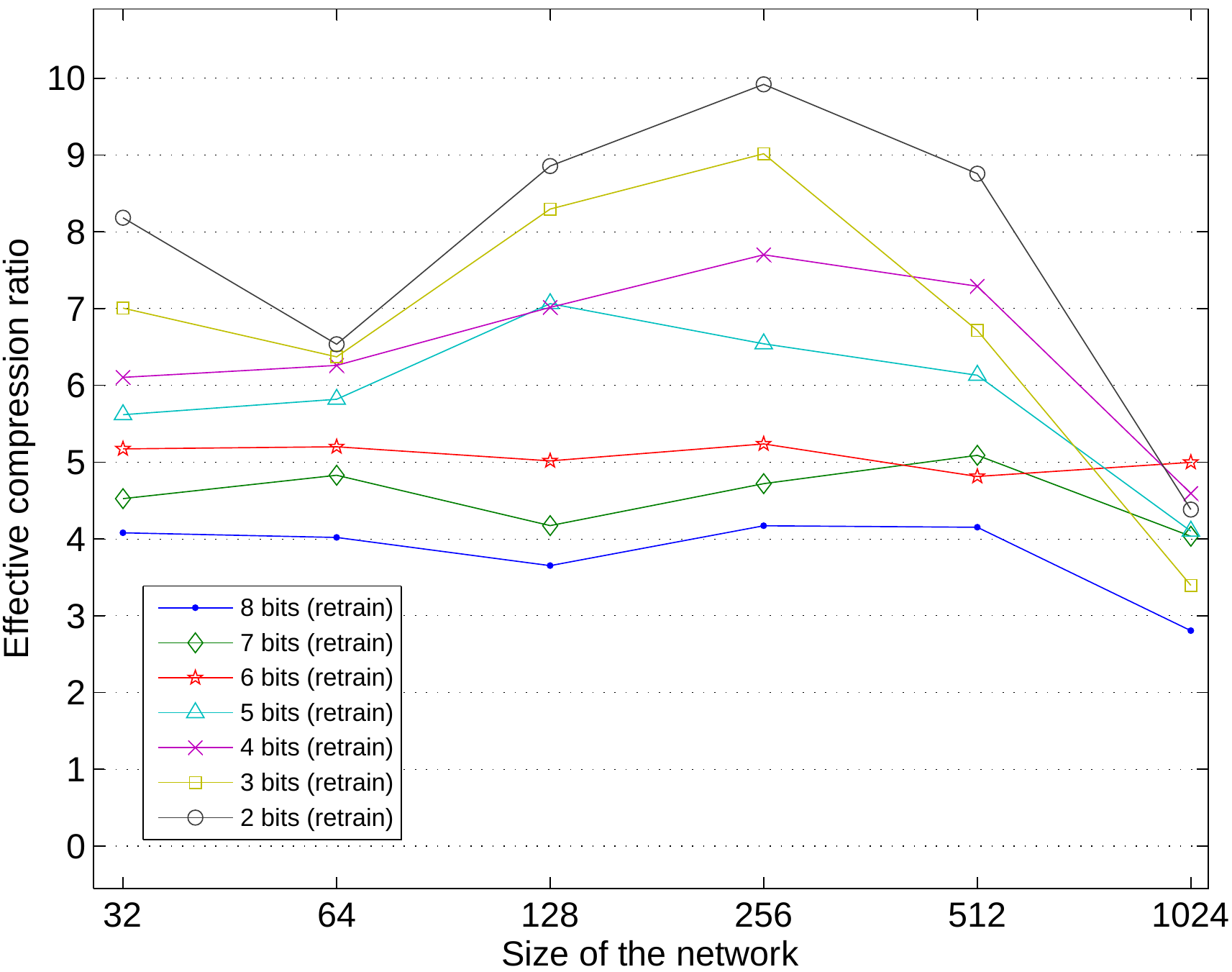}\label{fig_ecr_retrain}}
\caption{Effective compression ratio (ECR) with respect to the layer size and the number of bits per 
weights for (a) direct quantization and (b) retrain-based quantization.}
\label{fig_ecr}
\end{figure}

The remaining question is how much memory space can be saved by quantization while maintaining the accuracy. To examine this, we introduce a metric called \emph{effective compression ratio (ECR)}, which is defined as follows:
\begin{align}
ECR = \frac{\text{Effective uncompressed size}}{\text{Compressed size}} \label{eq_ecr}
\end{align}
The compressed size is the total memory bits required for storing all weights with quantization. The effective uncompressed size is the total memory size with 32-bit floating point representation when the network achieves the same accuracy as that of the quantized network.

\figurename~\ref{fig_ecr_lines} describes how to obtain the effective number of parameters for uncompressed networks. Specifically, by varying the size, we find the number of total parameters of the floating-point network that shows the same accuracy as the quantized one. After that, the effective uncompressed size can be computed by multiplying 32 bits to the effective number of parameters.

Once we get the corresponding effective uncompressed size for the specific network size and the number of quantization bits, the ECR can be computed by \eqref{eq_ecr}. The ECRs for the direct and retrain-based quantization for various network sizes and quantization bits are shown in \figurename~\ref{fig_ecr}. For the direct quantization, 5 bit quantization shows the best ECR except for the layer size of 1024. On the other hand, even 2 bit quantization performs better than the others after retraining. That is, after retraining, a bigger network with extreme ternary (2 bit) quantization is more efficient in terms of the memory usage for weights than any other smaller networks with higher quantization bits when they are compared at the same accuracy.

\section{Discussion}
\label{sec:discussion}

In this study, we control the network size by changing the number of units in the hidden layers, the number of feature maps, or the number of levels.
At any case, reduced complexity lowers the resiliency to quantization. 
We are now conducting similar experiments to the recurrent neural networks that are known to be more sensitive to quantization \citep{shin2015fixed}. 
This work seems to be directly related to several network optimization methods, such as pruning, fault tolerance, and decomposition \citep{yu2012exploiting,han2015deep,xue2013restructuring,rigamonti2013learning}.
In the pruning, retraining of weights is conducted after zeroing small valued weights.  
The effects of pruning, fault tolerance, and network decomposition efficiency would be dependent on the redundant representation capability of DNNs. 

This study can be applied to hardware efficient DNN design. 
For design with limited hardware resources, when the size of the reference DNN is relatively small, it is advised to employ a very low-precision arithmetic and, instead, increase the network complexity as much as the hardware capacity allows.
But, when the DNNs are in the performance saturation region, this strategy does not always gain much because growing the `already-big' network size brings almost no performance advantages. This can be observed in \figurename~\ref{fig_number_of_bits_b} and \figurename~\ref{fig_ecr_retrain} where 6 bit quantization performed best at the largest layer size (1,024).

\section{Conclusion}
\label{sec:conclusion}
We analyze the performance of fixed-point deep neural networks, an FFDNN for phoneme recognition and a CNN for image classification, while not only changing the arithmetic precision but also varying their network complexity.
The low-precision networks for this analysis are obtained by using the retrain based quantization method, and the network complexity is controlled by changing the configurations of the hidden layers or feature maps.
The performance gap between the floating-point and the fixed-point neural networks with ternary weights (+1, 0, -1) almost vanishes when the DNNs are in the performance saturation region for the given training data. 
However, when the complexity of DNNs are reduced, by lowering either the number of units, feature maps, or hidden layers, the performance gap between them increases. 
In other words, a large size network that may contain redundant representation capability for the given training data does not hurt by the lowered precision, but a very compact network does. 

\subsubsection*{Acknowledgments}

This work was supported in part by the Brain Korea 21 Plus Project and the National Research Foundation of Korea (NRF) grants funded by the Korea government (MSIP) (No. 2015R1A2A1A10056051).

\bibliography{iclr2016_conference}
\bibliographystyle{iclr2016_conference}

\end{document}